\documentclass[11pt,a4paper]{article}
\usepackage[hyperref]{acl2018}
\usepackage{times}
\usepackage{latexsym}
\usepackage{url}
\usepackage{amsmath}
\usepackage{amsfonts}
\usepackage{multirow}
\usepackage{microtype}
\usepackage{graphicx}
\usepackage{arydshln}
\usepackage{booktabs}
\usepackage{dsfont}

\usepackage{url}

\aclfinalcopy 

\title{Fully Statistical Neural Belief Tracking}
  
\author{Nikola Mrk\v{s}i\'{c}$^{\mathbf{1}}$ \and {Ivan Vuli\'c}$^{\mathbf{1,2}}$ \\
$^{\mathbf{1}}$ PolyAI \\
$^{\mathbf{2}}$ Language Technology Lab, University of Cambridge\\
\texttt{nikola@poly-ai.com} \hspace{1.5em} \texttt{ivan@poly-ai.com}\\
}

\date{}

\begin{document}
\maketitle
\begin{abstract}

This paper proposes an improvement to the existing data-driven Neural Belief Tracking (NBT) framework for Dialogue State Tracking (DST). The existing NBT model uses a hand-crafted belief state update mechanism which involves an expensive manual retuning step whenever the model is deployed to a new dialogue domain. We show that this update mechanism can be \textit{learned} jointly with the semantic decoding and context modelling parts of the NBT model, eliminating the last rule-based module from this DST framework. We propose two different statistical update mechanisms and show that dialogue dynamics can be modelled with a very small number of additional model parameters. In our DST evaluation over three languages, we show that this model achieves competitive performance and provides a robust framework for building resource-light DST models.

\end{abstract}

\section{Introduction}
\label{s:intro}
The problem of language understanding permeates the deployment of statistical dialogue systems. These systems rely on dialogue state tracking (DST) modules to model the user's intent at any point of an ongoing conversation \cite{young:10}. In turn, DST models rely on domain-specific Spoken Language Understanding (SLU) modules to extract turn-level user goals, which are then incorporated into the \emph{belief state}, the system's internal probability distribution over possible dialogue states. 

The dialogue states are defined by the domain-specific \emph{ontology}: it enumerates the constraints the users can express using a collection of slots (e.g.~\emph{price range}) and their slot values (e.g.~\emph{cheap, expensive} for the aforementioned slots). The belief state is used by the downstream dialogue management component to choose the next system response \cite{Su:16,Su:2017}.

A large number of DST models \cite[\textit{inter alia}]{Wang:13,Sun:16,Liu:2017,Vodolan:2017} treat SLU as a separate problem: the detached SLU modules are a dependency for such systems as they require large amounts of annotated training data. Moreover, recent research has demonstrated that systems which treat SLU and DST as a single problem have proven superior to those which decouple them \cite{Williams:16}. Delexicalisation-based models, such as the one proposed by \cite{Henderson:14d,Henderson:14b} offer unparalleled generalisation capability. 

These models use \emph{exact matching} to replace occurrences of slot names and values with generic tags, allowing them to share parameters across all slot values. This allows them to deal with slot values not seen during training. However, their downside is shifting the problem of dealing with linguistic variation back to the system designers, who have to craft \emph{semantic lexicons} to specify rephrasings for ontology values. Examples of such rephrasings are [\emph{cheaper}, \emph{affordable}, \emph{cheaply}] for slot-value pair \textsc{food=cheap}, or [\emph{with internet}, \emph{has internet}] for \textsc{has internet=True}. The use of such lexicons has a profound effect on DST performance \cite{Mrksic:16}. Moreover, such lexicons introduce a design barrier for deploying these models to large real-world dialogue domains and other languages.  



\begin{figure*} 
\begin{center}
\includegraphics[width = 0.94\textwidth]{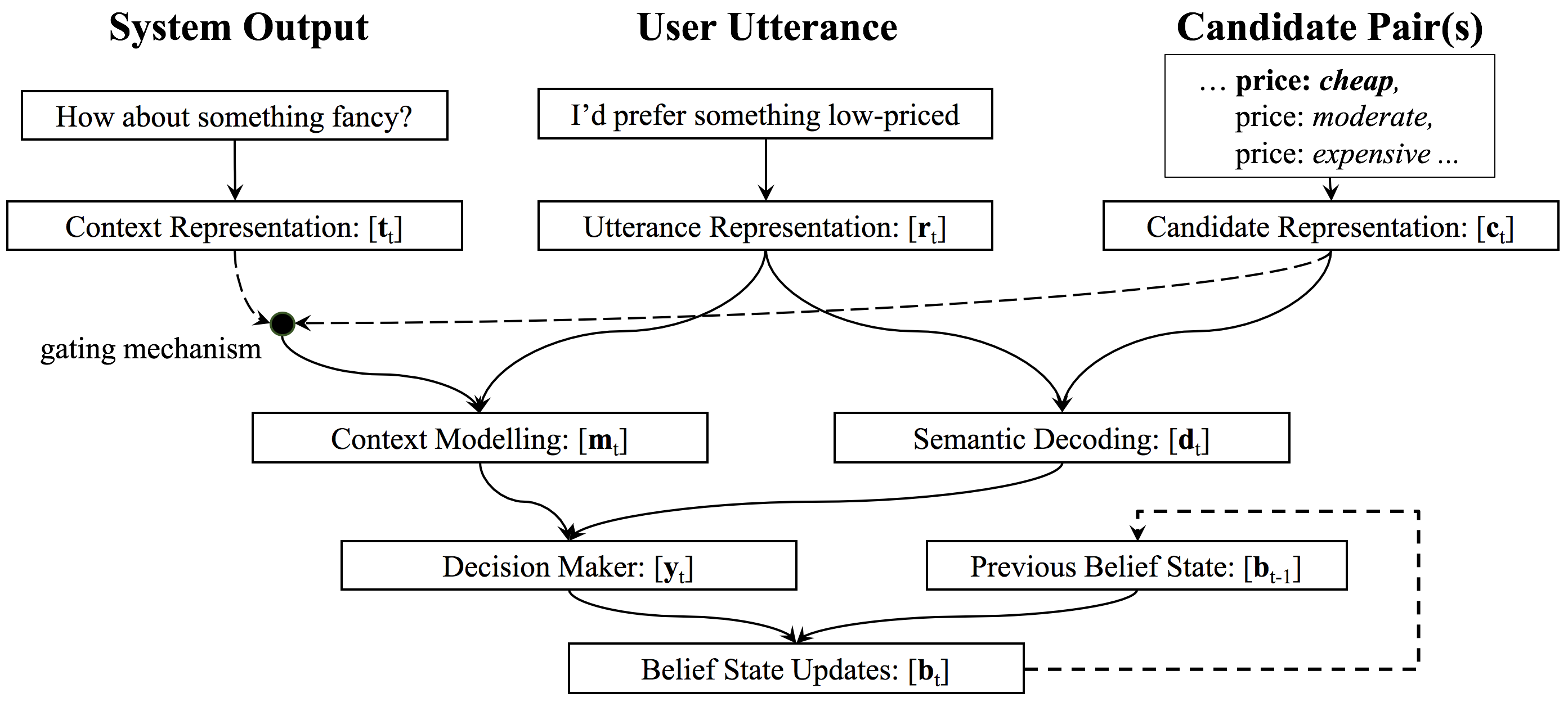}
\caption{The architecture of the fully statistical neural belief tracker. Belief state updates are not rule-based but learned jointly with the semantic decoding and context modelling parts of the NBT model.\label{fig:sys_diagram} }
\end{center}
\end{figure*}

The Neural Belief Tracker (NBT) framework \cite{Mrksic:17a} is a recent attempt to overcome these obstacles by using dense word embeddings in place of traditional $n$-gram features. By making use of semantic relations embedded in the vector spaces, the NBT achieves DST performance competitive to lexicon-supplemented delexicalisation-based models without relying on any hand-crafted resources. Moreover, the NBT framework enables deployment and bootstrapping of DST models for languages other than English \cite{Mrksic:17}. As shown by \newcite{Vulic:2017}, phenomena such as morphology make DST a substantially harder problem in linguistically richer languages such as Italian and German.  

The NBT models decompose the (per-slot) multi-class value prediction problem into many binary ones: they iterate through all slot values defined by the ontology and decide which ones have just been expressed by the user. To differentiate between slots, they take as input the word vector of the slot value that it is making a decision about. In doing that, the previous belief state is discarded. However, the previous state may contain pertinent information for making the turn-level decision. 

\paragraph{Contribution}
In this work, we show that cross-turn dependencies can be learned automatically: this eliminates the rule-based NBT component and effectively yields a fully statistical dialogue state tracker. Our competitive results on the benchmarking WOZ dataset for three languages indicate that the proposed fully statistical model: \textbf{1)}  is robust with respect to the input vector space, and \textbf{2)} is easily portable and applicable to different languages.

Finally, we make the code of the novel NBT framework publicly available at: \textit{https://github.com/nmrksic/neural-belief-tracker}, in hope of helping researchers to overcome the initial high-cost barrier to using DST as a real-world language understanding task.


\section{Methodology}
\label{s:methodology}
\paragraph{Neural Belief Tracker: Overview} The NBT models are implemented as multi-layer neural networks. Their input consists of three components: \textbf{1)} the list of vectors for words in the last user utterance; \textbf{2)} the word vectors of the slot name and value (e.g.~\textsc{food=Indian}) that the model is currently making a decision about; and \textbf{3)} the word vectors which represent arguments of the preceding system acts.\footnote{Following \newcite{Mrksic:17a}, we also consider only \emph{system requests} and \emph{system confirmations}, which ask the user to specify the value of a given slot (e.g. \emph{`What kind of venue are you looking for?'}), or to confirm whether a certain intent is part of their belief state (\emph{`Are you looking for Chinese food?'}).} To perform belief state tracking, the NBT model iterates over all candidate slot-value pairs as defined by the ontology, and decides which ones have just been expressed by the user.

The first layer of the NBT (see Figure~\ref{fig:sys_diagram}) learns to map these inputs into intermediate distributed representations of:  \textbf{1)} the current utterance representation $\mathbf{r}$; \textbf{2)} the current candidate slot-value pair $\mathbf{c}$; and \textbf{3)} the preceding system act $\mathbf{m}$. These representations then interact through the \emph{context modelling} and \emph{semantic decoding} downstream components, and are finally coalesced into the decision about the current slot value pair by the final \emph{binary decision making} module. For full details of this setup, see the original NBT paper \cite{Mrksic:17a}.

\subsection{Statistical Belief State Updates}
The NBT framework effectively recasts the per-slot multi-class value prediction problem as multiple binary ones: this enables the model to deal with slot values unseen in the training data. It iterates through all slot values and decides which ones have just been expressed by the user. 

In the original NBT framework \cite{Mrksic:17a}, the model for turn-level prediction is trained using SGD, maximising the accuracy of turn-level slot-value predictions. These predictions take preceding system acts into account, but \textit{not} the previous belief state. Note that these predictions are done \textit{separately for each slot value}.


\paragraph{Problem Definition} 
For any given slot $s \in V_s$, let $\mathbf{b}_{s}^{t-1}$ be the true belief state at time $t-1$ (this is a vector of length $|V_{s}|+2$, accounting for all slot values and two special values, \emph{dontcare} and \emph{NONE}). At turn $t$, let the intermediate representations representing the preceding system acts and the current user utterance be $\mathbf{m}^{t}$ and $\mathbf{r}^{t}$. If the model is currently deciding about slot value $v \in V_{s}$, let the intermediate candidate slot-value representation be $\mathbf{c}_{v}^{t}$. The NBT binary-decision making module produces an estimate $\mathbf{y}_{s,v}^{t} = P(s, v | \mathbf{r^{t}, m^{t}})$. We aim to combine this estimate with the previous belief state estimate for the entire slot $s$, ${\mathbf{b}_{s}}^{t-1}$, so that:

\vspace{-0.0mm}
\begin{equation}
\mathbf{b}_{s}^{t} = \phi(\mathbf{y}_{s}^{t}, \mathbf{b}_{s}^{t-1})
\label{eq:1}
\end{equation}
\noindent where $\mathbf{y}_{s}^{t}$ is the vector of probabilities for each of the slot values $v \in V_{s}$. 


\paragraph{Previously: Rule-Based} The original NBT framework employs a convoluted programmatic rule-based update which is hand-crafted and cannot be optimised or learned with gradient descent methods. For each slot value pair $(s,v)$, its new probability $\mathbf{b}_{s,v}^t$ is computed as follows:
\begin{equation}
\mathbf{b}_{s,v}^{t} = \lambda \mathbf{y}_{s,v}^{t} + (1-\lambda) \mathbf{b}_{s,v}^{t-1}
\label{eq:simple}
\end{equation}

\noindent $\lambda$ is a tunable coefficient which determines the
relative weight of the turn-level and previous turns'
belief state estimates, and is maximised according to DST performance on a validation set. For slot $s$, the set of its \textit{detected values} at turn $t$ is then given as follows:

\vspace{-0.0mm}
\begin{equation}
V_s^t = \{v \in V_s | \mathbf{b}_{s,v}^{t} \geq 0.5 \}
\label{eq:logic}
\end{equation}
For informables (i.e., goal-tracking slots), which unlike requestable slots require belief tracking across turns, if $V_s^t \neq \emptyset$ the value in $V_s^t$ with the highest probability is selected as the current goal.

This effectively means that the value with the highest probabilities $\mathbf{b}_{s,v}^{t}$ at turn $t$ is then chosen as the new goal value, but \textit{only} if its new probability $\mathbf{b}_{s,v}^{t}$ is greater than 0.5. If no value has probability greater than 0.5, the predicted goal value stays the same as the one predicted in the previous turn - even if its probability $\mathbf{b}_{s,v}^{t}$ is now less than 0.5.

In the rule-based method, tuning the hyper-parameter $\lambda$ adjusts how likely any predicted value is to override previously predicted values. However, the ``belief state'' produced in this manner is not a valid probability distribution. It just predicts the top value using an ad-hoc rule that was empirically verified by \newcite{Mrksic:17a}.\footnote{We have also experimented with a simple model which tunes the hyper-parameter $\lambda$ during training without the remaining decision logic at each turn $t$ (see Eq.~\eqref{eq:logic}). The belief state update is performed as follows: $\mathbf{b}_{s}^{t} = \lambda \mathbf{y}_{s}^{t} + (1-\lambda) \mathbf{b}_{s}^{t-1}$. We note that this simplistic update mechanism performs poorly in practice, with joint goal accuracy on the English DST task in the 0.22-0.25 interval (compare it to the results from Table~\ref{tab:en}).}


This rule-based approach comes at a cost: the NBT framework with such updates is little more than an SLU decoder capable of modelling the preceding system acts. Its parameters do not learn to handle the previous belief state, which is essential for probabilistic modelling in POMDP-based dialogue systems \cite{young:10c,Thomson:10}. We now show two update mechanisms that extend the NBT framework to (learn to) perform statistical belief state updates. 





\paragraph{1. One-Step Markovian Update} 
To stay in line with the NBT paradigm, the criteria for the belief state update mechanism $\phi$ from Eq.~\eqref{eq:1} are: \textbf{1)} it is a differentiable function that can be backpropagated during NBT training; and \textbf{2)} it produces a valid probability distribution  $\mathbf{b}_{s}^{t}$ as output. Figure~\ref{fig:sys_diagram} shows our fully statistical NBT architecture. 

The first learned statistical update mechanism, termed One-Step Markovian Update, combines the previous belief state $\mathbf{b}_{s}^{t-1}$ and the current turn-level estimate $\mathbf{y}_{s}^{t}$ using a one-step belief state update:

\begin{equation}
\mathbf{b}_{s}^{t} ~=~ softmax \left( W_{curr} \mathbf{y}_{s}^{t} + W_{past} \mathbf{b}_{s}^{t-1} \right)
\end{equation}
\noindent $W_{curr}$ and $W_{past}$ are matrices which learn to combine the two signals into a new belief state. This variant violates the NBT design paradigm: each row of the two matrices learns to operate over \emph{specific} slot values.\footnote{This means the model will not learn to predict or maintain slot values as part of the belief state if it has not encountered these values during training.} Even though turn-level NBT output $\mathbf{y}_{s}^{t}$ may contain the right prediction, the parameters of the corresponding row in $W_{curr}$ will not be trained to update the belief state, since its parameters (for the given value) will not have been updated during training. Similarly, the same row in $W_{past}$ will not learn to maintain the given slot value as part of the belief state. 

To overcome the data sparsity and preserve the NBT model's ability to deal with unseen values, one can use the fact that there are fundamentally only two different actions that a belief tracker needs to perform: \textbf{1)} maintain the same prediction as in the previous turn; or \textbf{2)} update the prediction given strong indication that a new slot value has been expressed. To facilitate transfer learning, the second update variant introduces additional constraints for the one-step belief state update.

\paragraph{2. Constrained Markovian Update} 
This variant constrains the two matrices so that each of them contains only two different scalar values. The first one populates the diagonal elements, and the other one is replicated for all off-diagonal elements:

\begin{align}
W_{curr, i, j} &= 
\begin{cases}
    a_{curr},& \text{if } i= j\\
    b_{curr},   & \text{otherwise}
\end{cases} \\
W_{past, i, j} &= 
\begin{cases}
    a_{past},& \text{if } i = j \\
    b_{past},   & \text{otherwise}
\end{cases}
\end{align}

\noindent where the four scalar values are learned jointly with other NBT parameters. The diagonal values learn the relative importance of propagating the previous value ($a_{past}$), or of accepting a newly detected value ($a_{curr}$). The off-diagonal elements learn how turn-level signals ($b_{curr}$) or past probabilities for other values ($b_{past}$) impact the predictions for the current belief state. The parameters acting over all slot values are in this way tied, ensuring that the model can deal with slot values unseen in training. 



\section{Experimental Setup}
\label{s:exp}


\paragraph{Evaluation: Data and Metrics}
As in prior work the DST evaluation is based on the Wizard-of-Oz (WOZ) v2.0 dataset \cite{Wen:17,Mrksic:17a}, comprising 1,200 dialogues split into training (600 dialogues), validation (200), and test data (400). The English data were translated to German and Italian by professional translators \cite{Mrksic:17}. In all experiments, we report the standard DST performance measure: \textit{joint goal accuracy}, which is defined as the proportion of dialogue turns where all the user's search goal constraints were correctly identified. Finally, all reported scores are averages over 5 NBT training runs.

\paragraph{Training Setup} We compare three belief state update mechanisms (rule-based vs. two statistical ones) fixing all other NBT components as suggested by \newcite{Mrksic:17a}: the better-performing NBT-CNN variant is used, trained by Adam \cite{adam:15} with dropout \cite{Srivastava:2014} of 0.5, gradient clipping, batch size of 256, and 400 epochs. All model hyperparameters were tuned on the validation sets.

\begin{table}[t]
\centering
\def\arraystretch{1.0}
{\small
\begin{tabular}{l cc}
\toprule
{} &  \multicolumn{2}{c}{ \bf English WOZ 2.0} \\ 
{\bf Model Variant} & \textsc{glove} (\textsc{dist}) & \textsc{paragram-sl999}  \\
\cmidrule(lr){2-3}
\bf Rule-Based & 80.1   &  84.2 \\
\cmidrule(lr){2-3}
\bf  1. One-Step & 80.8  & 82.1  \\ 
\bf  2. Constrained  &  \textbf{81.8} & \textbf{84.8}  \\  
\bottomrule
\end{tabular}}%
\caption{The English DST performance (\textit{joint goal accuracy}) with standard input word vectors (\S\ref{s:exp}).} 
\label{tab:en}
\end{table}

\paragraph{Word Vectors} 
To test the model's robustness, we use a variety of standard word vectors from prior work. For English, following \newcite{Mrksic:17a} we use \textbf{1)} distributional \textsc{glove} vectors \cite{Pennington:14}, and \textbf{2)} specialised \textsc{paragram-sl999} vectors \cite{Wieting:15}, obtained by injecting similarity constraints from the Paraphrase Database \cite{Pavlick:2015acl} into \textsc{glove}.


For Italian and German, we compare to the work of \newcite{Vulic:2017}, who report state-of-the-art DST scores on the Italian and German WOZ 2.0 datasets. In this experiment, we train the models using distributional skip-gram vectors with a large vocabulary (labelled \textsc{dist} in Table~\ref{tab:it-de}). Subsequently, we compare them to models trained using  word vectors specialised using similarity constraints derived from language-specific morphological rules (labelled \textsc{spec} in Table~\ref{tab:it-de}).








\section{Results and Discussion}
\label{s:results}

Table~\ref{tab:en} compares the two variants of the statistical update. The Constrained Markovian Update is the better of the two learned updates, despite using only four parameters to model dialogue dynamics (rather than $O(V^2)$, $V$ being the slot value count). This shows that the ability to generalise to unseen slot values matters more than the ability to model value-specific behaviour. In fact, combining the two updates led to no performance gains over the stand-alone Constrained Markovian update. 

Table~\ref{tab:it-de} investigates the portability of this model to other languages. The statistical update shows comparable performance to the rule-based one, outperforming it in three out of four experiments. In fact, our model trained using the specialised word vectors sets the new state-of-the-art performance for English, Italian and German WOZ 2.0 datasets.  This supports our claim that eliminating the hand-tuned rule-based update makes the NBT model more stable and better suited to deployment across different dialogue domains and languages.

\begin{table}[t]
\centering
{\small
\begin{tabular}{l  cc|cc}
\toprule
{} &  \multicolumn{2}{c|}{ \bf Italian} & \multicolumn{2}{c}{ \bf German }\\ 
{} &  \bf {\textsc{dist}} & \bf {\textsc{spec}} & \bf {\textsc{dist}} & \bf {\textsc{spec}} \\ 
\bf Rule-Based Update & \textbf{74.2}  & 76.0 & 60.6 & 66.3 \\
\bf  Learned Update &  73.7 &  \textbf{76.1} &  \textbf{61.5} & \textbf{68.1} \\
\bottomrule
\end{tabular}}%
\caption{DST performance on Italian and German. Only results with the better scoring learned Constrained Markovian Update are reported. } 
\label{tab:it-de}
\end{table}

\paragraph{DST as Downstream Evaluation} All of the experiments show that the use of semantically specialised vectors benefits DST performance. The scale of these gains is robust across all experiments, regardless of language or the employed belief state update mechanism. So far, it has been hard to use the DST task as a proxy for measuring the correlation between word vectors' intrinsic performance (in tasks like SimLex-999 \cite{Hill:2014}) and their usefulness for downstream language understanding tasks. Having eliminated the rule-based update from the NBT model, we make our evaluation framework publicly available in hope that DST performance can serve as a useful tool for measuring the correlation between intrinsic and extrinsic performance of word vector collections.  

\section{Conclusion}
This paper proposed an extension to the Neural Belief Tracking (NBT) model for Dialogue State Tracking (DST) \cite{Mrksic:17a}. In the previous NBT model, system designers have to tune the \emph{belief state update} mechanism manually whenever the model is deployed to new dialogue domains. On the other hand, the proposed model \emph{learns} to update the belief state automatically, relying on no domain-specific validation sets to optimise DST performance. Our model outperforms the existing NBT model, setting the new state-of-the-art-performance for the Multilingual WOZ 2.0 dataset across all three languages. We make the proposed framework publicly available in hope of providing a robust tool for exploring the DST task for the wider NLP community.

\section*{Acknowledgments}
We thank the anonymous reviewers for their helpful and insightful suggestions. We are also grateful to Diarmuid \'{O} S\'{e}aghdha and Steve Young for many fruitful discussions.


\bibliographystyle{acl_natbib}
\bibliography{nbt2_refs}

\end{document}